\documentclass[conference]{IEEEtran}
\IEEEoverridecommandlockouts
\usepackage{cite}
\usepackage{amsmath,amssymb,amsfonts}
\usepackage{algorithmic}
\usepackage{graphicx}
\usepackage{textcomp}
\usepackage{eurosym}
\usepackage{tikz}
\usetikzlibrary{shapes}
\usepackage{xcolor}
\def\BibTeX{{\rm B\kern-.05em{\sc i\kern-.025em b}\kern-.08em
    T\kern-.1667em\lower.7ex\hbox{E}\kern-.125emX}}

\newcommand{\refequation}[1]{\eqref{#1}}
\newcommand{\reffigure}[1]{\figurename~\ref{#1}}
\newcommand{\refsection}[1]{Section~\ref{#1}}
\newcommand{\refsubsection}[1]{Subsection~\ref{#1}}

\newcommand{\transpose}[1]{#1^\top}

\newcommand{\argmax}{\operatornamewithlimits{argmax}}
    
\begin{document}

\title{Biasing MCTS with Features for General Games\\
\thanks{Funded by a \euro2m ERC Consolidator Grant (http://ludeme.eu).}
}

\author{\IEEEauthorblockN{Dennis J. N. J. Soemers, {\'E}ric Piette, and Cameron Browne}
\IEEEauthorblockA{\textit{Department of Data Science and Knowledge Engineering} \\
\textit{Maastricht University}\\
Maastricht, the Netherlands \\
\texttt{\{dennis.soemers,eric.piette,cameron.browne\}@maastrichtuniversity.nl}}
}

\maketitle

\begin{abstract}
This paper proposes using a linear function approximator, rather than a deep neural network (DNN), to bias a Monte Carlo tree search (MCTS) player for general games. This is unlikely to match the potential raw playing strength of DNNs, but has advantages in terms of generality, interpretability and resources (time and hardware) required for training. Features describing local patterns are used as inputs. The features are formulated in such a way that they are easily interpretable and applicable to a wide range of general games, and might encode simple local strategies. We gradually create new features during the same self-play training process used to learn feature weights. We evaluate the playing strength of an MCTS player biased by learnt features against a standard upper confidence bounds for trees (UCT) player in multiple different board games, and demonstrate significantly improved playing strength in the majority of them after a small number of self-play training games.
\end{abstract}


\section{Introduction} \label{Sec:Introduction}

Combinations of search algorithms with learning from self-play have led to strong results in game-playing AI for various board games \cite{Silver2017AlphaGoZero,Anthony2017ExIt,Silver2018AlphaZero} and video games \cite{Jiang2018FeedbackBased}. 
Currently, a common combination is to use {\it Monte Carlo tree search} (MCTS) \cite{Kocsis2006UCT,Coulom2007,Browne2012} for search and {\it deep neural networks} (DNNs) for learning.

While DNN-based learning approaches have advanced the state of the art in terms of raw playing strength for game-playing AI, they have a number of disadvantages in comparison to other learning techniques %(such as simple linear function approximators) 
in other aspects. For example, effectively training the large DNNs used to obtain state-of-the-art results in board games \cite{Silver2017AlphaGoZero,Anthony2017ExIt,Silver2018AlphaZero} requires significant training time and/or large amounts of hardware. 
Different games typically require different DNN architectures -- the input and output layers in particular are game-specific -- and separate training processes starting from scratch per game. Due to the large number of trainable parameters, and the use of low-level inputs (i.e. raw board states), it is often difficult to extract interpretable knowledge such as strategies from a DNN during or after training.

In this paper, we describe and evaluate an approach to simultaneously grow a set of features, and learn weights for a linear policy function using those features, from self-play. 
Each feature \cite{Browne2019StrategicFeatures} describes a simple local pattern, and is specified in a general manner that is applicable to many different games and easily interpretable. 
%An added bonus is that each feature may encode simple local strategies relevant to the games being modelled, potentially giving some insight into their strategic potential.

The focus of this work is not on achieving state-of-the-art game-playing performance. 
Our main contribution is to demonstrate that, using a simple linear policy function, our learned features and weights can improve the performance of a standard MCTS player across a variety of board games given a set of severe restrictions:
\begin{itemize}
\item Features are specified in a general format, compatible with many different games \cite{Browne2019StrategicFeatures}.
\item We do not manually construct game-specific feature sets.
\item We use a low amount of training time; up to 300 games of self-play per game, with 5 seconds of thinking time allowed per move.
\item We use relatively little hardware (every sequence of self-play runs sequentially on a single node).
\item The learned function must be able to work correctly for any number of legal actions (i.e., we do not provide an upper bound on the size of the action space, as is typically required for the output layer in DNN-based approaches).
\item Hyperparameters (for MCTS as well as the self-play learning process) have not been optimised (their values are manually selected), and the same hyperparameter values are used across all games.
\end{itemize}

Under the restrictions listed above, we find that the performance of MCTS can already be improved relatively easily in multiple different games. 
We expect that the interpretability of the features, in combination with a simple linear function, could provide insight into what strategies are relevant to the games being modelled, potentially giving some insight into their strategic potential and the relationships between different games in terms of strategy \cite{Browne2019StrategicFeatures}. 

The generality of the approach may also create opportunities for transferring of strategies between games, which combined with the low computational requirements could allow the approach to be applied to large numbers of games and variants within reasonable times. 
This is crucial for the Digital Ludeme Project \cite{Browne2018ModernTechniques}, which involves digitally modelling large numbers of ancient games and exploring relationships between them.

\refsection{Sec:Background} provides relevant background information for this paper. The self-play training process used to learn weights for a linear function is described in \refsection{Sec:ExpertIteration}. Our approach for automatically constructing and growing the set of features per game is explained in \refsection{Sec:GrowingFeatureSet}. \refsection{Sec:Experiments} discusses the experimental evaluation of the approach. The paper is concluded in \refsection{Sec:Conclusion}.

\section{Background} \label{Sec:Background}

This section provides background information on the \textsc{Ludii} general game system \cite{Browne2018ModernTechniques}, which we use to run different games used for evaluation, and the format in which we specify general game features \cite{Browne2019StrategicFeatures}.

\subsection{The \textsc{Ludii} General Game System}

The \textsc{Ludii} {\it general game system} \cite{Browne2018ModernTechniques} implements units of game-related information, referred to as \textit{ludemes} \cite{Parlett2016Ludeme}, in different Java classes. A single ludeme can, for example, be a simple game rule that describes a particular kind of legal move, or it can be a description of a component (a piece or a board) used to play a game. A complete game can be described as a tree of ludemes. For example, \reffigure{Fig:LudiiTicTacToe} depicts how the game of Tic-Tac-Toe can be modelled in \textsc{Ludii}.

\begin{figure}
\centering
\fbox{%
  \parbox{.97\columnwidth}{%
  	\texttt{%
  		(game "Tic-Tac-Toe"\\
  			\hspace*{0.5em}(play \{ (player "P1")(player "P2") \})\\
  			\hspace*{0.5em}(equipment\\
  				\hspace*{1em}\{ (board "Board" (square 3)) \}\\
  				\hspace*{1em}\{ (disc "Piece") (cross "Cross")) \}\\
  			\hspace*{0.5em})\\
  			\hspace*{0.5em}(rules\\
  				\hspace*{1em}(moves (to Mover (empty)))\\
  				\hspace*{1em}(end (line length:3)(result Mover win))\\
  			\hspace*{0.5em})\\
  		)
  	}%
  }%
}
\caption{The game of Tic-Tac-Toe modelled in \textsc{Ludii}.}
\label{Fig:LudiiTicTacToe}
\end{figure}

\subsection{Specification of Features} \label{Subsec:FeatureSpecification}

In {\it general game playing} (GGP) research based on the Stanford {\it game definition language} (GDL) \cite{Genesereth2005GGP}, various approaches have been proposed for using general game features \cite{Kirci2011GGPFeatureLearning, Waledzik2011GGPEvalFunctions, Michulke2012DistanceFeaturesGGP, Michulke2013AdmissibleDistanceHeuristics, Waledzik2014AutoGenEvalFuncGGP}. 
These are all built around the logic-based formalism of GDL, and therefore not directly applicable to general game systems that use a different language (such as \textsc{Ludii}).

Following \cite{Browne2019StrategicFeatures}, the features $x$ used in this paper consist of:
\begin{enumerate}
\item A \textit{pattern} $p_x$, which contains one or more elements that the feature tests for in relative positions. The different types of elements that a feature can test for in different positions are \textit{off-board}, \textit{empty}, \textit{friendly piece}, \textit{enemy piece}, \textit{piece owned by player $n$} (for any $n$), and \textit{piece with unique index $n$} (in the game's definition).
\item A description of an \textit{action} $a_x$ which the feature ``recommends'' playing (in practice a feature may also discourage playing its action if a negative weight is learned for that feature). The action $a_x$ is specified by two relative positions; a position to move ``from'' and a position to move ``to''. For games like Hex and Go, only the ``to'' position is relevant. For games such as Draughts and Chess, also the ``from'' position is relevant.
\end{enumerate}

\begin{figure}
\centering
\includegraphics[width=1\linewidth]{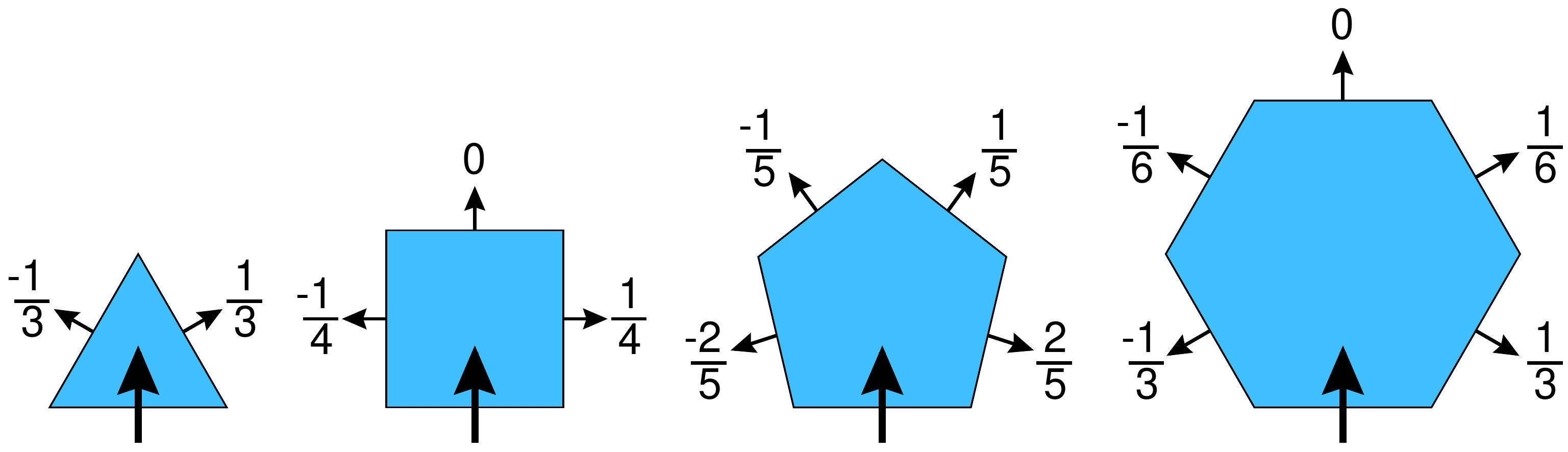}
\caption{Fractional steps and their corresponding relative steps.}
\label{Fig:TurnsDifferentCells}
\end{figure}

Relative positions in patterns and action specifications of features are described as sequences of numbers, referred to as \textit{walks}. The length of such a sequence corresponds to the number of steps to take through a graph representation of the playable area (typically a board), starting from some fixed \textit{anchor position}. Every number in the sequence denotes how far we should rotate (as a fraction of a full $360^\circ$, clockwise turn), relative to the ``current'' direction, before taking the next step. \reffigure{Fig:TurnsDifferentCells} depicts different fractions and their corresponding movement directions in different types of cells, relative to a ``current'' direction pointing to the north.

\begin{figure}[htbp]
\centering
\includegraphics[width=.85\linewidth]{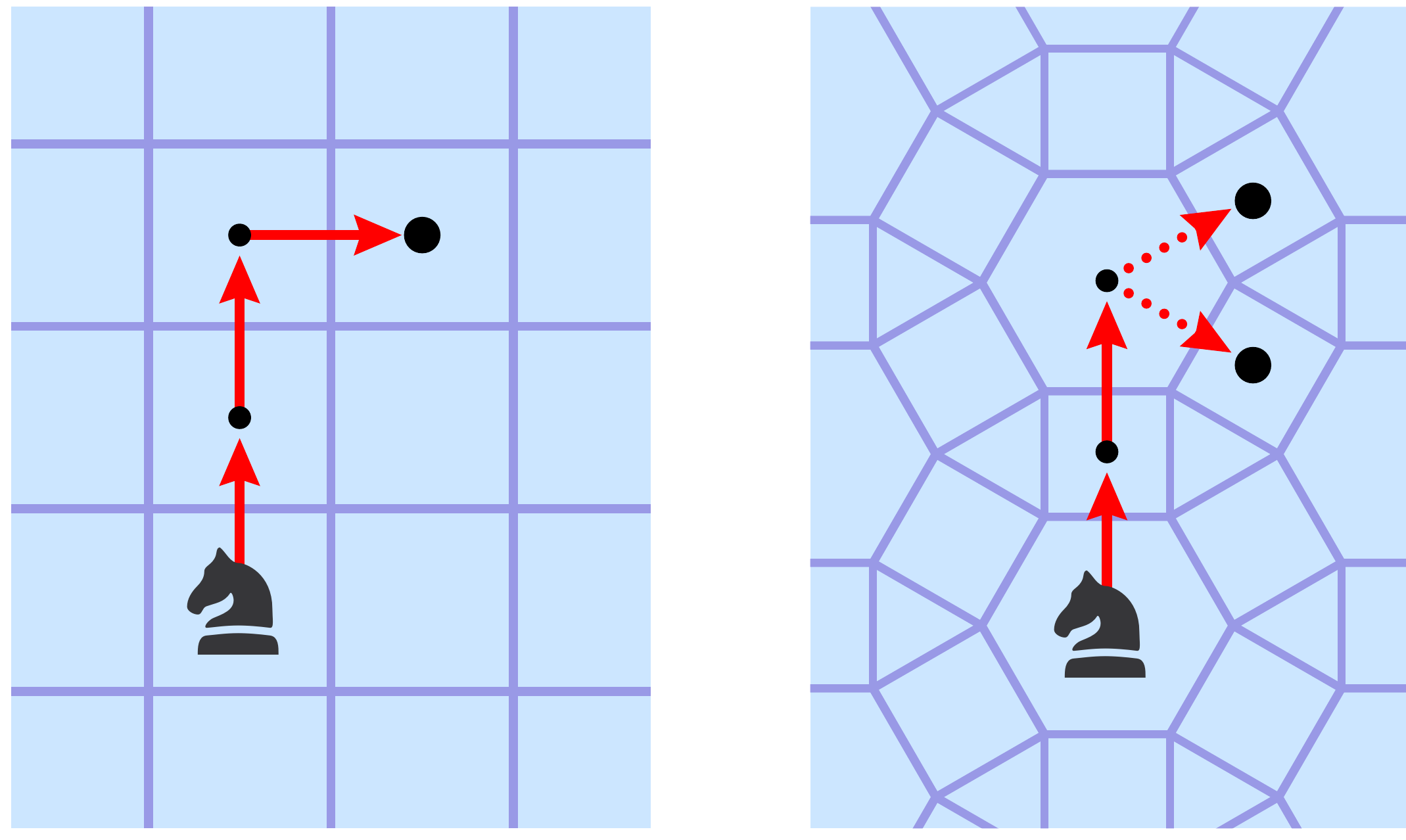}
\caption{Relative position(s) reached by a $\{ 0, 0, \frac{1}{4} \}$ walk in two different boards. In the board on the right-hand side, the $\frac{1}{4}$ turn can be rounded to a turn of $\frac{1}{3}$ or $\frac{1}{6}$.}
\label{Fig:KnightSteps}
\end{figure}

Features defined in this manner are applicable to any game that can be modelled as being played on some graph, where vertices may contain pieces that may be owned by players. This applies to many board games, and potentially also games without an explicit board. The \textsc{Ludii} system uses such a graph to model the playable area of any game. Every vertex contains a list of references to adjacent vertices, sorted in a consistent manner to facilitate indexing based on clockwise turns, with null entries to facilitate off-board checks. \reffigure{Fig:KnightSteps} depicts how a $\{ 0, 0, \frac{1}{4} \}$ walk can specify relative positions with a similar semantic meaning in two different types of boards.

\section{Expert Iteration with a Linear Policy} \label{Sec:ExpertIteration}

In the {\it expert iteration} framework \cite{Anthony2017ExIt, Silver2017AlphaGoZero}, an \textit{apprentice} policy and an \textit{expert} policy are used to iteratively improve each other during self-play. The apprentice policy is a trainable, computationally efficient component, such as a DNN. Given any state $s$, it can compute a distribution $\boldsymbol{p}(s)$ over the set of actions $A(s)$ that are legal in $s$ in a fixed amount of time. The expert is generally a component that involves more ``deliberation'' (i.e., search or planning), such as MCTS.

The main idea of expert iteration is to view the expert policy as a \textit{policy improvement operator} for the apprentice. 
The apprentice can be used to bias the searching behaviour of the expert, and the expert can use additional computation time to adjust (ideally improve) the distribution computed by the apprentice. 
The adjusted distribution can subsequently be used as a learning target by the apprentice.

\subsection{Formalisation of the Apprentice}

The particular combination of a DNN as apprentice, and MCTS as expert, has led to state-of-the-art game-playing performance in various board games \cite{Silver2017AlphaGoZero, Anthony2017ExIt, Silver2018AlphaZero}, but the DNNs have a number of important drawbacks in aspects other than raw playing strength, as listed in \refsection{Sec:Introduction}. Therefore we investigate using a linear function rather than a DNN in this paper. We aim to train a function of the form given by \refequation{Eq:LinearFunction}:
\begin{equation} \label{Eq:LinearFunction}
f(s, a) = \transpose{\boldsymbol{\theta}} \boldsymbol{\phi}(s, a),
\end{equation}
where $\boldsymbol{\theta}$ is a vector of trainable parameters, $\boldsymbol{\phi}(s, a)$ is a (binary) feature vector for a state-action pair $(s, a)$, and $f(s, a) \in \mathbb{R}$ is a real-valued output for the same state-action pair. Given a set of legal actions $A(s)$ in a state $s$, the complete distribution $\boldsymbol{p}(s)$ over all actions $a_i \in A(s)$ is computed by applying the softmax function to a vector of outputs $f(s, a_i)$:
\begin{equation} \label{Eq:Softmax}
p_i(s) = \frac{\exp(f(s, a_i))}{\sum_{k = 1}^{\vert A(s) \vert} \exp(f(s, a_k))}.
\end{equation}

When DNNs are used as apprentice, it is customary to have an output layer with one output node per unique action that may ever be legal in any given game state. This can easily lead to excessively large numbers of outputs in some games, such as $11,259$ outputs in Shogi \cite{Silver2018AlphaZero}. It also requires domain knowledge in the form of an accurate upper bound on the number of unique actions, which is a problem in terms of generality. The number of outputs computed in any given state $s$ by \refequation{Eq:Softmax} is equal to the number of legal actions $\vert A(s) \vert$ in that state, which is typically multiple orders of magnitude lower in a game like Shogi \cite{Iida2002ComputerShogi}. Of course, an advantage of DNNs may be that its computational requirements remain constant regardless of $\vert A(s) \vert$, whereas the computational requirements of \refequation{Eq:Softmax} scale linearly with the number of legal actions $\vert A(s) \vert$.

\subsection{Formalisation of Feature Vectors}

Suppose that we have some set of features $\mathcal{X}$, where every feature $x \in \mathcal{X}$ is specified as described in \refsubsection{Subsec:FeatureSpecification}. 
More details on how such a feature set is created will be provided in the next section. We say that a feature $x$ is \textit{active} for a state-action pair $(s, a)$ if there exists some \textit{anchor} position in the game's underlying graph such that, after applying any necessary rotation and/or reflection:
\begin{enumerate}
\item The feature's action $a_x$ corresponds to the action $a$.
\item All elements of the feature's pattern $p_x$ are satisfied in the game state $s$.
\end{enumerate}
Relative positions in $a_x$ and $p_x$ are evaluated by resolving the walks, starting from the anchor position.

We define state-action feature vectors $\boldsymbol{\phi}(s, a)$ as binary vectors that contain a value of $1$ for features $x$ that are active for the state-action pair $(s, a)$, and a value of $0$ for all other features. Note that different \textit{instantiations} (with different anchor positions, rotations, or reflections) of the same feature may be active simultaneously; in such a case, we still simply assign a feature value of $1$.

\subsection{Guiding the Expert using the Apprentice}

We use a learned apprentice policy to guide the expert (MCTS) in its \textit{selection} and \textit{play-out} phases. The most common selection strategy \cite{Kocsis2006UCT} is to follow the UCB1 policy \cite{Auer2002FiniteTimeMAB}. Given a current node with a state $s$, it selects the child node corresponding to the action $a_{ucb1}$ given by \refequation{Eq:UCB1}.
\begin{equation} \label{Eq:UCB1}
a_{ucb1} = \argmax_a \hat{Q}(s, a) + C_{ucb1} \sqrt{\frac{\ln \left( \sum_{a'} N(s, a') \right)}{N(s, a)}}.
\end{equation}
$\hat{Q}(s, a)$ denotes the estimated value of playing $a$ in $s$ based on previous MCTS iterations (i.e. the average score backpropagated through the node reached by executing $a$ in $s$), $C_{UCB1}$ is a hyperparameter (the ``exploration constant''), and $N(s, a)$ denotes the \textit{visit count} of $(s, a)$ (the number of previous MCTS iterations that have selected $a$ in the current node). The sum $\sum_{a'} N(s, a')$ is equivalent to the total number of previous MCTS iterations that have reached the current node. This strategy only uses statistics gathered by MCTS itself, and does not use the apprentice.

In this paper, we use the apprentice to guide the selection step of MCTS using the same strategy as AlphaGo Zero \cite{Silver2017AlphaGoZero}, which selects the action $a_{puct}$ given by \refequation{Eq:PUCT}.
\begin{equation} \label{Eq:PUCT}
a_{puct} = \argmax_a \hat{Q}(s, a) + C_{puct} p(s, a) \frac{\sqrt{\sum_{a'} N(s, a')}}{1 + N(s, a)}.
\end{equation}
$C_{puct}$ is a hyperparameter, and $p(s, a)$ denotes the probability of selecting $a$ in $s$ according to the distribution computed by the apprentice.

The most straightforward approach for using the apprentice in the play-out step is to select actions according to the probability distribution computed by the apprentice, rather than selecting actions uniformly at random. The computational overhead of computing feature vectors can be mitigated by transitioning from using the apprentice policy early in play-outs to a uniform random policy in later parts of play-outs.

\subsection{Training Apprentice with Expert Iteration}

We train the apprentice policy in a similar way to \cite{Silver2017AlphaGoZero}, interpreting the visit counts at the end of an MCTS search process as the target distribution. Let $N(s, a)$ denote the number of MCTS iterations that selected action $a$ in state $s$. For any node in the search tree with a state $s$, the quantity $\frac{N(s, a)}{\sum_{a'} N(s, a')}$ can then be interpreted as the probability assigned to $a$ in state $s$ by the MCTS expert policy.

Let $\boldsymbol{\pi}(s)$ denote a vector of such probabilities for all legal actions $a \in A(s)$, and let $\boldsymbol{p}(s)$ denote a similar vector computed by the apprentice policy. The loss function is then given by \refequation{Eq:LossFunction}:
\begin{equation} \label{Eq:LossFunction}
\mathcal{L}(s) = - \transpose{\boldsymbol{\pi}(s)} \log \boldsymbol{p}(s) + \frac{\lambda}{2} \left\lVert \boldsymbol{\theta} \right\rVert^2,
\end{equation}
which computes the cross-entropy loss between the two distributions, and an $L_2$ regularisation penalty with a hyperparameter $\lambda$. A stochastic gradient descent (SGD) update to reduce this loss, based on a single example $s$, can be implemented according to \refequation{Eq:SgdUpdate}:
\begin{equation} \label{Eq:SgdUpdate}
\boldsymbol{\theta} \gets \boldsymbol{\theta} - \alpha \sum_{a \in A(s)} \left[ \left( p(s, a) - \pi(s, a) \right) \times \boldsymbol{\phi}(s, a) \right] - \alpha \lambda \boldsymbol{\theta},
\end{equation}
where $p(s, a)$ and $\pi(s, a)$ denote the entries corresponding to action $a$ in the $\boldsymbol{p}(s)$ and $\boldsymbol{\pi}(s)$ vectors, respectively, and $\alpha$ is a step-size hyperparameter.

\section{Growing Feature Set During Self-Play} \label{Sec:GrowingFeatureSet}

In prior research using similar types of features to those used in this paper for game-playing AI applications, the complete set of features to use is typically determined before any parameters for policies or value functions using those features are learned. 

For example, large sets of features were exhaustively generated in the games of Go \cite{Gelly2007CombiningOnlineOffline}, Hearts \cite{Sturtevant2007FeatureConstructionHearts}, and Breakthrough \cite{Lorentz2017PatternsBreakthrough}. Such exhaustive sets can easily contain tens of thousands of features. When features are used in a pure Reinforcement Learning agent \cite{Sturtevant2007FeatureConstructionHearts}, or implemented for a single specific game \cite{Gelly2007CombiningOnlineOffline, Lorentz2017PatternsBreakthrough} which permits a highly efficient game-specific implementation (for instance by incrementally updating the set of active features \cite{Gelly2007CombiningOnlineOffline}), it is computationally tractable to use such large feature sets. In our general game system, we find that the computational overhead of computing active features already becomes detrimental to game-playing performance in some games for feature sets containing only hundreds of features (see \refsection{Sec:Experiments}).

Other approaches \cite{Skowronski2009AutomatedDiscovery} for discovering sets of useful features often consist of iteratively modifying feature sets (e.g., by adding new features in some manner), and evaluating the usefulness of such modifications by running a number of evaluation games using the features. This can be slow when many evaluation games are required for an accurate evaluation. In this section, we propose an approach that starts with a small set of initial features, and gradually adds more complex features during the self-play expert iteration process, guided by the same loss function used to train the policy.

\subsection{Initial Feature Set}

Many ludemes used by the \textsc{Ludii} system to describe legal moves can easily be extended with functionality to generate patterns $p_x$ or (parts of) features $x$ that detect legal moves. For example, the ``\texttt{(to Mover (empty))}'' ludeme, used in Tic-Tac-Toe (\reffigure{Fig:LudiiTicTacToe}) as well as many other games (such as Hex, Yavalath, etc.), can generate a feature $x$ which: 
\begin{itemize}
\item Recommends playing in the feature's \textit{anchor} position. 
\item Has a pattern $p_x$ that requires the same anchor position (specified using a $0$-length walk) to be empty.
\end{itemize}
Formally, we implement ludemes used in the specification of move rules to generate a feature set $\mathcal{X}$ such that, for any possible game state $s$ and legal action $a \in A(s)$, there exists at least one feature $x \in \mathcal{X}$ that is active for the state-action pair $(s, a)$. In the worst case (for highly complex movement rules), this can simply be an ``empty'' feature without any restrictions on either the action or the game state (i.e. a feature that is always active).

Such features are generally not interesting features by themselves. However, they can be used to reduce the space of candidate features considered for subsequent addition to the feature set. We only allow new features to be added to the feature set if they are at least as restrictive as, and not incompatible with, at least one of the features generated by movement rule ludemes. This enables us to automatically ignore many features that would be useless due to never being active in gameplay. For example, features that require an enemy or friendly piece in the location where they recommend playing will never be considered in games that only permit playing in empty positions.

For every feature generated by move rule ludemes as described above, we generate a set of ``atomic'' features $x$, which have exactly one requirement for an element specified in their pattern $p_x$ (in addition to any requirements that may already be there due to move rule ludemes). We exhaustively generate all such atomic features, with generated walks restricted to at most two steps. Using only atomic features (no features with more complex patterns) keeps the initial feature count down to a low number. We use the maximum number of adjacencies of any vertex in a game's graph to determine the number of potentially meaningful rotations in a game.

\subsection{Adding New Features During Expert Iteration}

In the expert iteration framework, experience generated from self-play is used to update the parameters $\boldsymbol{\theta}$ of the apprentice policy, such that its output distributions $\boldsymbol{p}$ more closely match the distributions $\boldsymbol{\pi}$ of the expert policy. The \textit{error} $p(s, a) - \pi(s, a)$, which also appears in the SGD update rule in \refequation{Eq:SgdUpdate}, has a large absolute value for state-action pairs $(s, a)$ for which the distributions do not yet closely match, and a low absolute value if the distributions already closely match. 

We propose to use this error value as an indicator of state-action pairs $(s, a)$ for which it is beneficial to add new features to the feature set. The intuition is that there is no need to add extra features for $(s, a)$ pairs for which $p(s, a)$ already closely matches $\pi(s, a)$, but extra features are more likely to be useful if they activate for $(s, a)$ pairs for which $p(s, a)$ and $\pi(s, a)$ do not yet closely match. 

Whenever we wish to add a new feature to the feature set, we sample a batch $E = \{ \left< s_i, A(s_i), \boldsymbol{\pi}_i (s_i) \right> \}$ of samples of experience collected from self-play. Every tuple $\left< s_i, A(s_i), \boldsymbol{\pi}_i (s_i) \right>$ in this batch contains a game state $s_i$ encountered in self-play, the list of legal actions $A(s_i)$ in that state, and the distribution $\boldsymbol{\pi}_i (s_i)$ over the actions $A(s_i)$ as computed by the expert at the point in time when this experience was saved. This batch is sampled without replacement from a larger experience buffer, in which we store exactly one new sample of experience (corresponding to the current game state) for every game state encountered during self-play.

Every pair of two features instances $(x_i, x_j)$ that are active together for at least one state-action pair across the entire batch $E$ is taken into consideration as a candidate pair that could be combined into a single new feature $x_i x_j$. Such a combination $x_i x_j$ is a new feature in which the patterns of the constituents $x_i$ and $x_j$ are merged. Note that a combined feature $x_i x_j$ will always be active for state-action pairs in which $x_i$ and $x_j$ were both active.

The candidate pair $(x_i, x_j)$ that maximises the score given by \refequation{Eq:CorrelationScore} is added to the feature set as a new feature.
\begin{equation} \label{Eq:CorrelationScore}
score(x_i, x_j) = \vert r_{err}(x_i, x_j) \vert \times \left( 1 - \vert r_{x_i x_j} (x_i, x_j) \vert \right)
\end{equation}
In this equation, $r_{err}(x_i, x_j)$ denotes the Pearson correlation coefficient between errors $p(s, a) - \pi(s, a)$, and the event of simultaneously observing features $x_i$ and $x_j$ to be active for a state-action pair $(s, a)$. Similarly, $r_{x_i x_j}(x_i, x_j)$ denotes the Pearson correlation coefficient between the event of simultaneously observing features $x_i$ and $x_j$ to be active, and the event of observing one of the constituents $x_i$ or $x_j$ to be active (whichever constituent leads to the strongest correlation is picked). Correlation coefficients are measured across all state-action pairs that occur in the complete batch $E$.

This score implements the intuition that new features are likely to be useful if they correlate strongly with observed errors between the apprentice and expert distributions, but are less likely to be useful if their activations correlate strongly with the activations of other features in the feature set. Similar intuition has also been shown to be useful for offline feature selection in supervised machine learning \cite{Hall1999CorrelationBasedFeatureSelection}. Ideally we would minimise correlations of candidate features between \textit{all} features in a feature set, but this is computationally expensive. Only computing correlations between candidate features and their constituents is significantly cheaper.

\section{Experiments} \label{Sec:Experiments}

This section describes a number of experiments which evaluate the effect of features and weights learned from self-play on the game-playing performance of MCTS in a variety of board games.

\subsection{Setup}

\begin{figure*}[t]
\centering
\includegraphics[width=.95\textwidth]{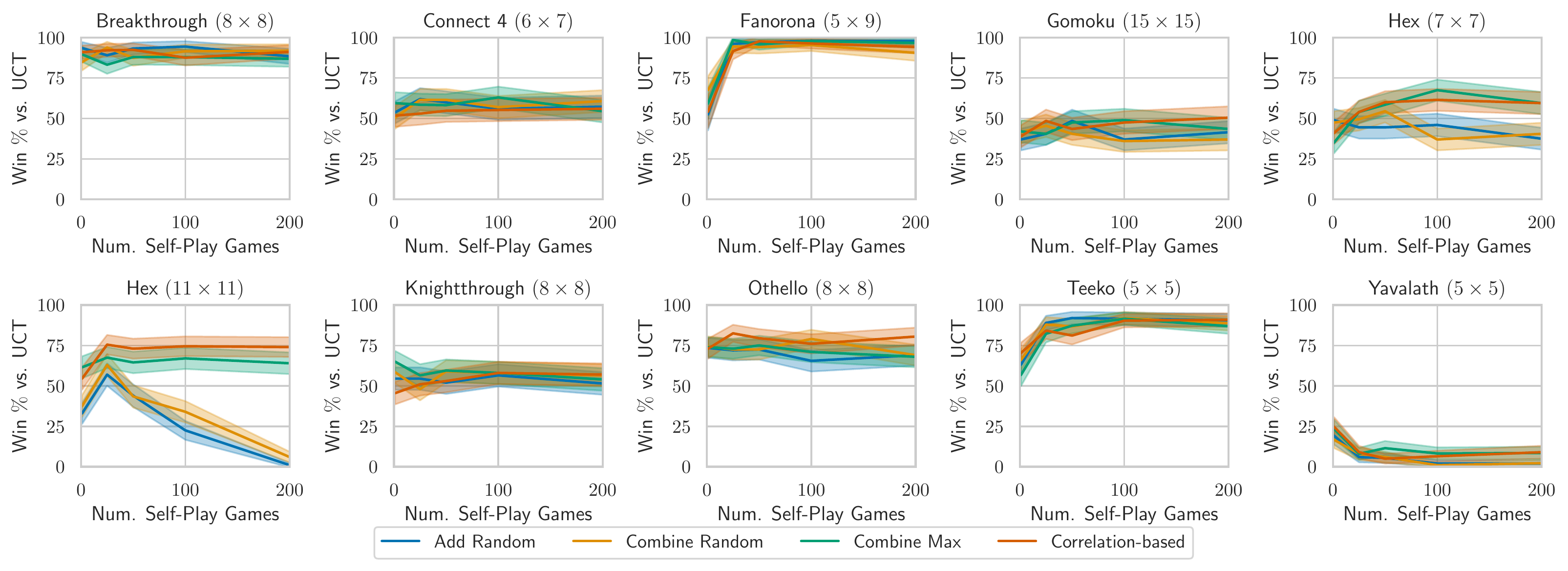}
\vspace{-8pt}
\caption{Learning curves for four different feature discovery strategies, over 200 games of self-play. Shaded regions depict $95\%$ confidence intervals for the win percentage of \textit{Biased MCTS} vs. \textit{UCT}. Performance evaluated by playing 200 evaluation games using feature sets and learned weights after $1$, $25$, $50$, $100$, and $200$ games of self-play.}
\vspace{-4pt}
\label{Fig:LearningCurves}
\end{figure*}

In the self-play process of expert iteration, experience is generated by equivalent MCTS agents playing against each other. They use \refequation{Eq:PUCT} in the selection phase, with $C_{puct} = \sqrt{2}$. The first move of every play-out is sampled from the apprentice distribution $\boldsymbol{p}$, and the corresponding node is added to the search tree. Subsequent play-out moves are selected uniformly at random, to avoid additional computational overhead of computing active features. Final moves for the ``real'' games are sampled from the expert distribution $\boldsymbol{\pi}$. We use $5$ seconds of ``thinking time'' per move. Games are terminated automatically after 100 moves, regardless of the game's standard rules.

We use an experience buffer with a maximum capacity of $200$ to store tuples of experience. Every move played in self-play results in one new tuple of experience. Old tuples are removed to make room for new tuples if necessary. We run one SGD update to update the apprentice parameters $\boldsymbol{\theta}$ after every move, with a step-size $\alpha = 0.05$, and $\lambda = 10^{-6}$ for $L_2$ regularisation. Gradients are computed and averaged across a batch of size $20$, sampled from the experience buffer.

We add one new feature after every game of self-play. A larger batch size of $30$ is used in this procedure. We evaluate three simpler feature discovery strategies, in addition to the correlation-based variant described in detail in \refsection{Sec:GrowingFeatureSet}:
\begin{enumerate}
\item \textit{Add Random}: This variant randomly selects pairs of simultaneously activated feature instances to combine. This can be viewed as an unguided baseline strategy.
\item \textit{Combine Random}: Randomly combines two feature instances that activate together in the state-action pair $(s, a)$ that maximises the absolute error $\vert p(s, a) - \pi(s, a) \vert$.
\item \textit{Combine Max}: Combines two feature instances that activate together in the state-action pair $(s, a)$ that maximises the absolute error $\vert p(s, a) - \pi(s, a) \vert$, such that one of them has the greatest absolute weight in the vector $\boldsymbol{\theta}$, and the other is selected randomly.
\item \textit{Correlation-based}: Combines feature instances such that \refequation{Eq:CorrelationScore} is maximised, as described in \refsection{Sec:GrowingFeatureSet}.
\end{enumerate}

We evaluate the performance of a \textit{Biased MCTS} agent (using features and weights learned from self-play) against a standard {\it upper confidence bounds for trees} (\textit{UCT}) agent:
\begin{itemize}
\item \textit{Biased MCTS}: Equal to the agent used during self-play, except that it selects moves with maximum visit counts rather than sampling moves from the $\boldsymbol{\pi}$ distribution during evaluation games.
\item \textit{UCT}: Uses \refequation{Eq:UCB1} in the selection phase, with $C_{ucb1} = \sqrt{2}$. Selects moves uniformly at random in play-outs. Plays moves with maximum visit counts in evaluation games.
\end{itemize}
All MCTS agents (in self-play as well as evaluation games) use relevant parts of the search tree built up when searching for previous moves to initialise the search tree for subsequent moves. Just like self-play games, evaluation games allow for 5 seconds of thinking time per move, and are automatically declared a tie after 100 moves.

We use nine different board games with standard board sizes, all implemented in the \textsc{Ludii} system, for evaluation. For the game of Hex, we use a $7 \times 7$ board in addition to the standard $11 \times 11$ board.

\subsection{Results - Growing Feature Set}

\begin{figure}
\centering
\includegraphics[width=.85\linewidth]{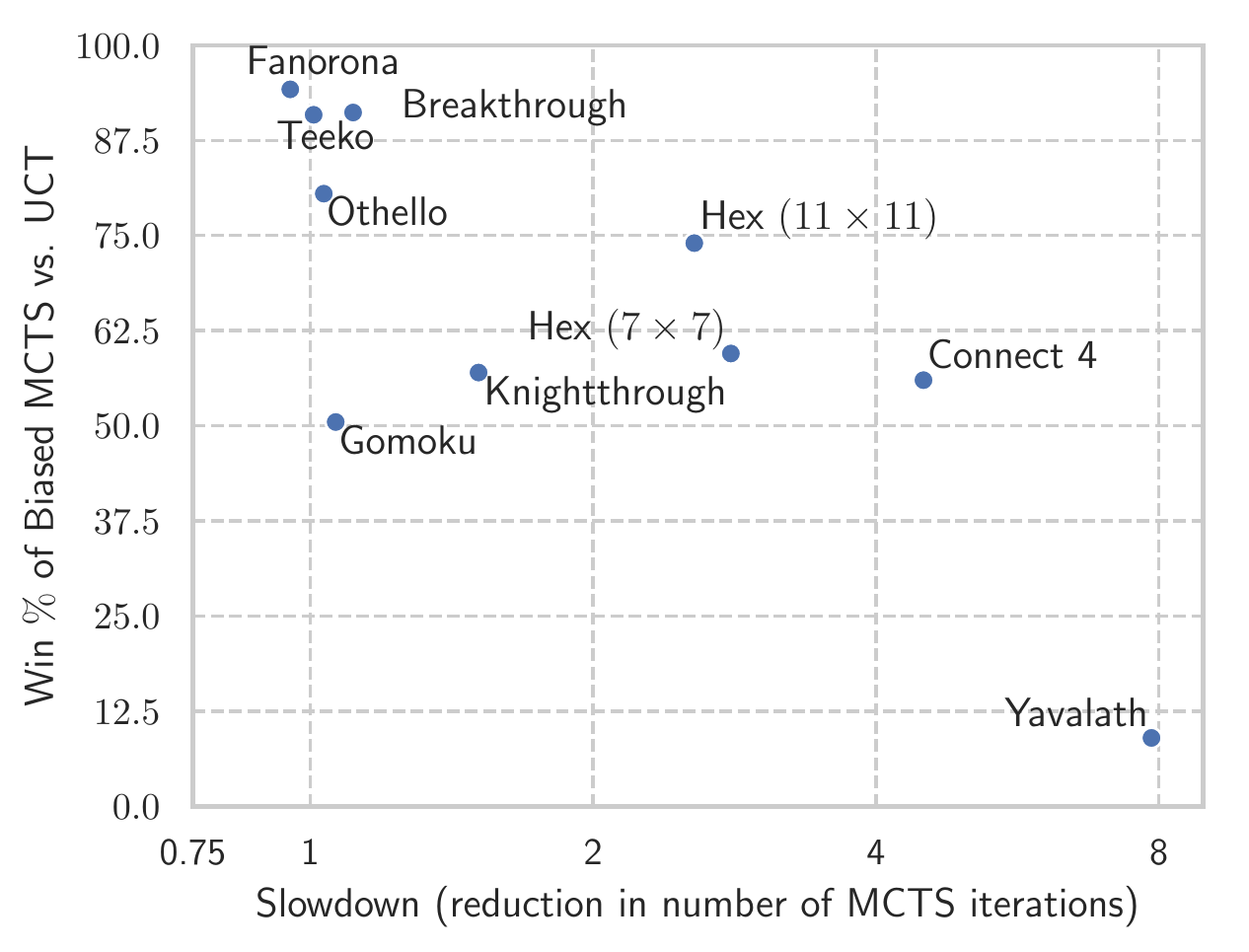}
\vspace{-8pt}
\caption{Relation between win percentage of \textit{Biased MCTS} vs. \textit{UCT} after 200 games of self-play, and the slowdown (reduction in MCTS iteration count) due to the computational overhead of using features.}
\label{Fig:ScoresVsSlowdowns}
\end{figure}

\begin{figure}
\centering
\includegraphics[width=.85\linewidth]{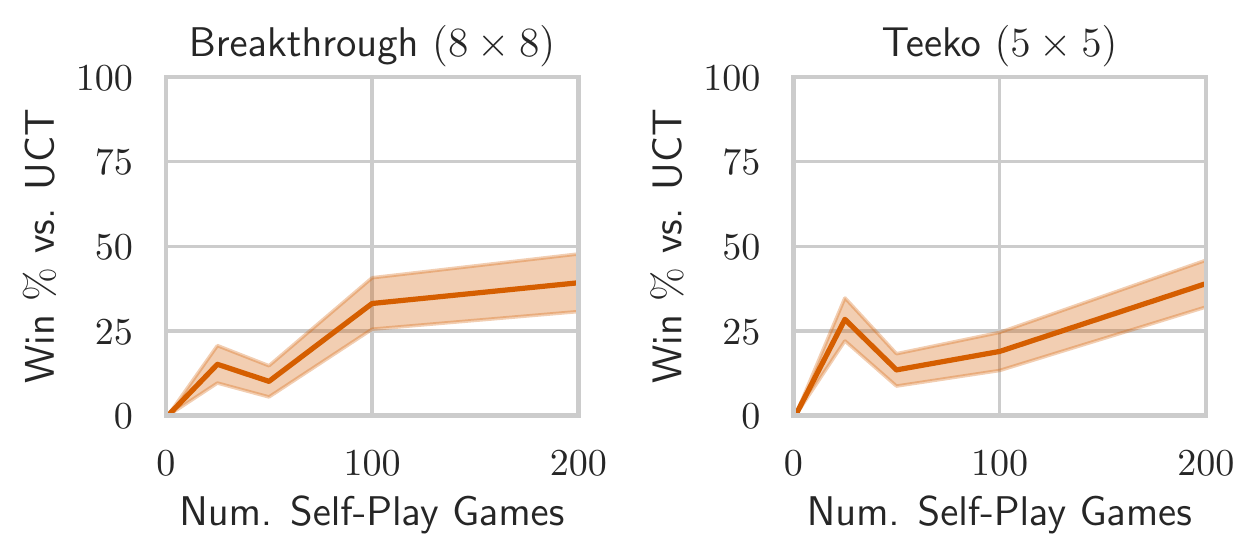}
\vspace{-8pt}
\caption{Learning curves for greedy linear agent (without any search) vs. \textit{UCT} in the games of Breakthrough and Teeko.}
\label{Fig:GreedyLinAgent}
\end{figure}

\begin{figure*}[t]
\centering
\includegraphics[width=.95\textwidth]{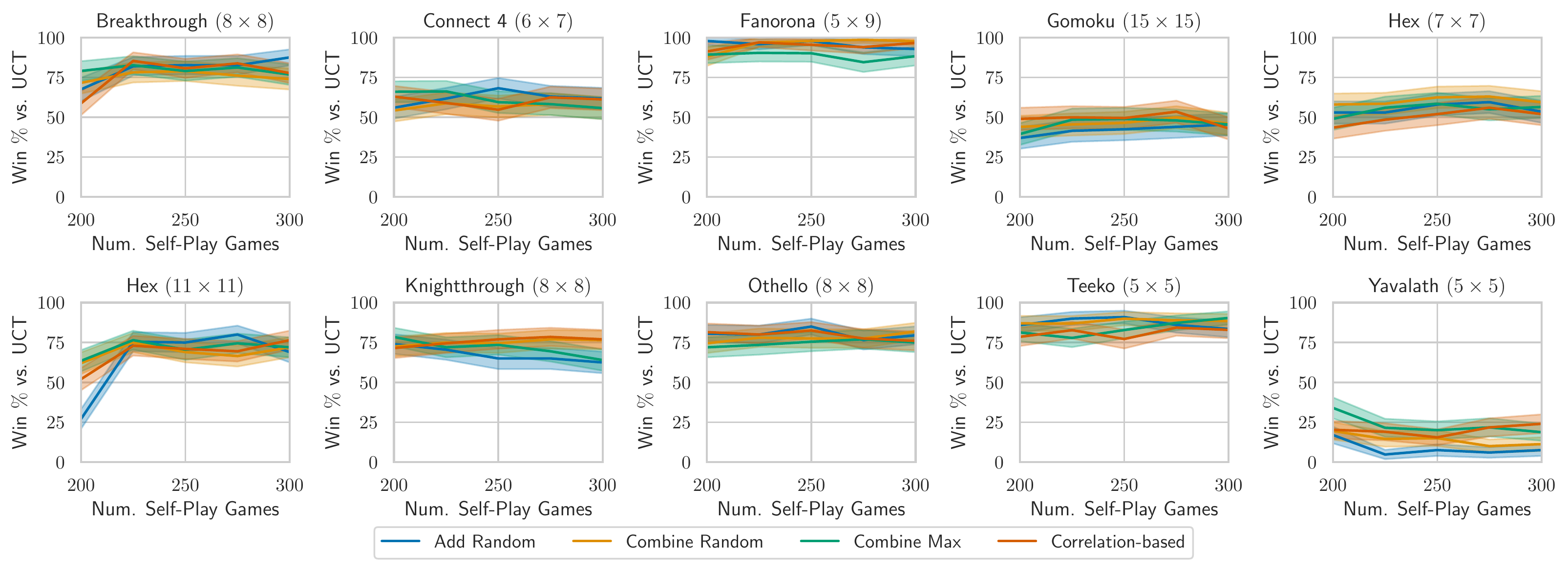}
\vspace{-8pt}
\caption{Learning curves for pruned features sets, over 100 additional games of self-play. Shaded regions depict $95\%$ confidence intervals for the win percentage of \textit{Biased MCTS} vs. \textit{UCT}. Performance evaluated by playing 200 evaluation games using weights learned after $0$, $25$, $50$, $75$, and $100$ games of self-play (after the first sequence of $200$ self-play games).}
\vspace{-4pt}
\label{Fig:LearningCurvesPruned}
\end{figure*}

\begin{figure}
\centering
\includegraphics[width=.85\linewidth]{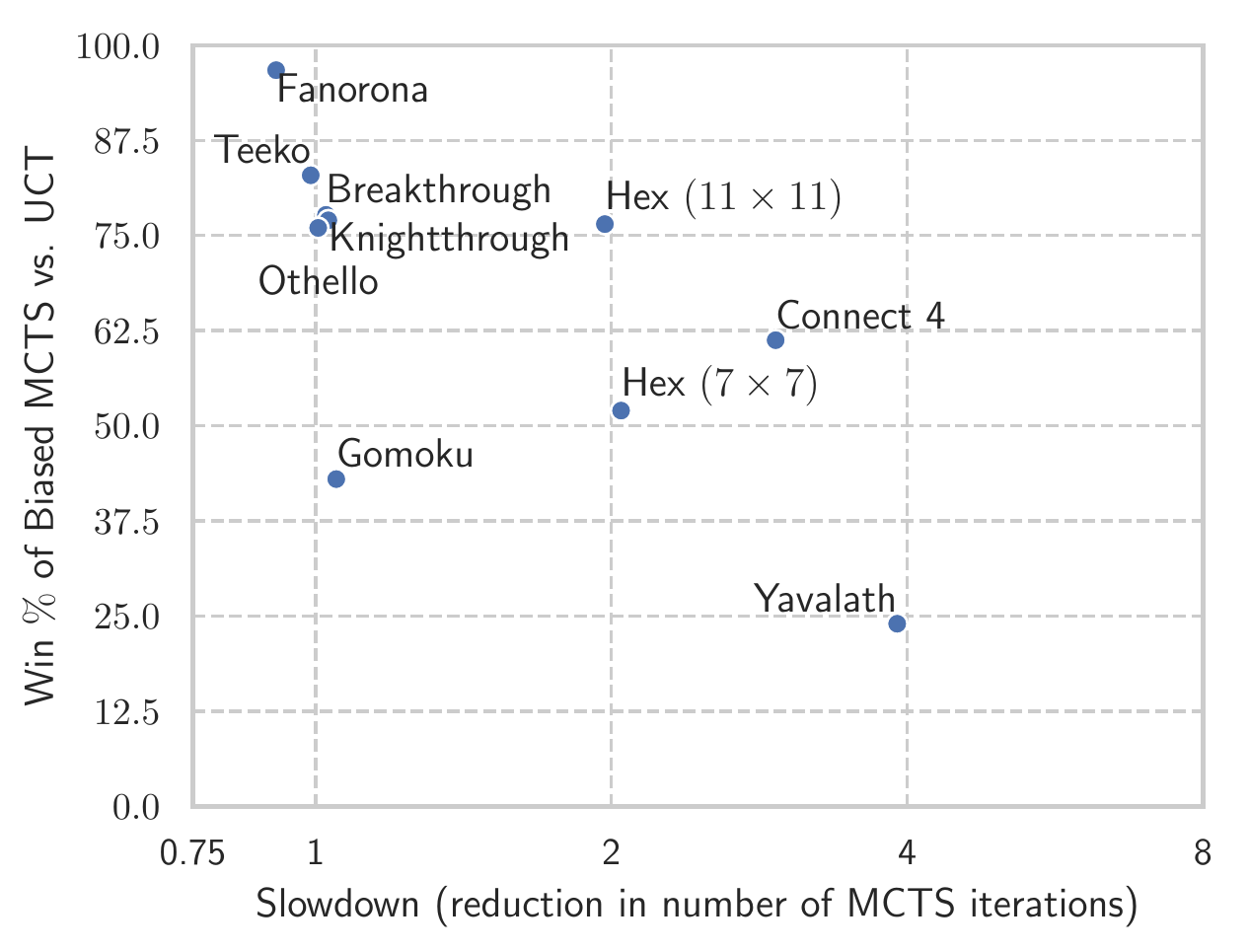}
\vspace{-8pt}
\caption{Relation between win percentage of \textit{Biased MCTS} vs. \textit{UCT} using pruned feature sets, and the slowdown (reduction in MCTS iteration count) due to the computational overhead of using features.}
\vspace{-4pt}
\label{Fig:ScoresVsSlowdownsPruned}
\end{figure}

\reffigure{Fig:LearningCurves} depicts learning curves for the four different feature discovery strategies, for all ten (variants of) games. At different checkpoints (after $1$, $25$, $50$, $100$, and $200$ games of self-play), we play 200 evaluation games where the \textit{Biased MCTS} agent plays against the benchmark \textit{UCT} agent, using the latest feature set and learned weights available at that checkpoint. \reffigure{Fig:LearningCurves} depicts $95\%$ confidence intervals for the win percentage of \textit{Biased MCTS} against \textit{UCT} at every checkpoint. Ties count as half a win for each player.

\textit{Biased MCTS} can quickly learn to outperform \textit{UCT} in the majority of games; by a significant margin in Breakthrough, Fanorona, Hex ($11 \times 11$), Othello, and Teeko, and by a smaller margin -- but often still statistically significant for the strongest variants -- in Connect 4, Hex ($7 \times 7$), and Knightthrough. Note that in some of these games there already is a significant gain in playing strength from just a single game of self-play, but in others there is a clear benefit in training for a longer time and growing the feature set. In Gomoku there is no apparant change in playing strength, and in Yavalath the playing strength is reduced by a significant margin. In most games the simpler feature discovery strategies already perform well, but we find the correlation-based feature discovery strategy to be better in some games, and never significantly worse.

\reffigure{Fig:ScoresVsSlowdowns} depicts the relation between the performance of \textit{Biased MCTS} after 200 games of self-play against UCT (with win percentage on the $y$-axis), and the slowdown due to computing active features (reduction in number of MCTS iterations on the $x$-axis). The slowdown per game is computed as $\frac{I_{uct}}{I_{biased}}$, where $I_{uct}$ and $I_{biased}$ denote the average number of complete MCTS iterations performed by \textit{UCT} and \textit{Biased MCTS}, respectively, in the first two moves per game. We only take into account the first two moves per game, because later moves can have wildly varying iteration counts depending on the game state. These results are given for the feature set learned using the \textit{Correlation-based} strategy. Computing features leads to the worst slowdown in Yavalath (an $8$ times reduction in MCTS iteration count), which may explain the poor performance in terms of win percentage in that game. In general, the computational overhead tends to be most noticeable in games for which the game implementation itself is highly efficient in \textsc{Ludii}, and hardly noticeable in games where the game logic itself requires more computation.

\reffigure{Fig:GreedyLinAgent} depicts learning curves for a greedy linear agent, which greedily plays actions $a \in A(s)$ such that $p(s, a)$ is maximised without performing any tree search at all, for the games of Breakthrough and Teeko. Surprisingly, we find that 200 games of self-play is already sufficient in these games to train a greedy agent that is competitive (reaching a win percentage of $40\%$) against UCT. Learning curves for other games (in which a simple greedy player still has a win percentage close to $0\%$ against UCT, as we would expect) are omitted to save space.

\subsection{Results - Pruned Feature Set}

To reduce the overhead of computing active features, we pruned the feature sets of all games by keeping only the $15$ features per game with the greatest absolute weights in the learned parameter vectors $\boldsymbol{\theta}$. Starting with the weights learned from the initial 200 games of self-play, we run 100 additional games of self-play to adjust the weights (for which different values may be better now that many other features have been pruned), but freeze the feature set. \reffigure{Fig:LearningCurvesPruned} depicts learning curves where performance is evaluated for $0$, $25$, $50$, $75$, and $100$ additional self-play games after pruning the feature set. Note that the four different feature discovery strategies may only have influence on the feature set and initial weights in this figure; they do not matter otherwise because no more features are added during these self-play games.

The relations between playing strength and slowdowns in MCTS iteration counts with pruned feature sets are depicted in \reffigure{Fig:ScoresVsSlowdownsPruned}. In comparison to the unpruned feature sets of \reffigure{Fig:ScoresVsSlowdowns}, we observe the most significant changes in performance for the games of Knightthrough (where \textit{Biased MCTS} now has a significant advantage over \textit{UCT}), and Yavalath (where slowdowns and win percentage for \textit{Biased MCTS} have clearly been improved, but playing strength is still worse than \textit{UCT}'s).

\subsection{Interpreting Features in Yavalath}

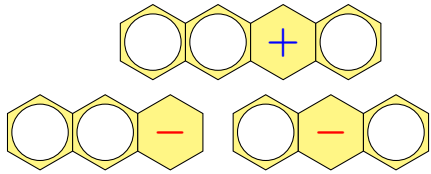
\begin{figure}[t]
\centering
\begin{tikzpicture} [hexa/.style= {shape=regular polygon,regular polygon sides=6,minimum size=1cm,draw,inner sep=0,fill=white!40!yellow,shape border rotate=30},
whitepiece/.style= {shape=circle,minimum size=0.75cm,draw,inner sep=0,fill=white}]

% first feature cells
\node at (0,0) [hexa] () {};
\node at ({sin(60)}, 0) [hexa] () {};
\node at ({2*sin(60)}, 0) [hexa] () {\Large $\color{blue}{\boldsymbol{+}}$};
\node at ({3*sin(60)}, 0) [hexa] () {};

% first feature pieces
\node at (0,0) [whitepiece] () {};
\node at ({sin(60)},0) [whitepiece] () {};
\node at ({3*sin(60)},0) [whitepiece] () {};

% second feature cells
\node at (-1.5,-1.2) [hexa] () {};
\node at ({-1.5 + sin(60)}, -1.2) [hexa] () {};
\node at ({-1.5 + 2*sin(60)}, -1.2) [hexa] () {\Large $\color{red}{\boldsymbol{-}}$};

% second feature pieces
\node at (-1.5,-1.2) [whitepiece] () {};
\node at ({-1.5 + sin(60)},-1.2) [whitepiece] () {};

% third feature cells
\node at (1.5,-1.2) [hexa] () {};
\node at ({1.5 + sin(60)}, -1.2) [hexa] () {\Large $\color{red}{\boldsymbol{-}}$};
\node at ({1.5 + 2*sin(60)}, -1.2) [hexa] () {};

% third feature pieces
\node at (1.5,-1.2) [whitepiece] () {};
\node at ({1.5 + 2*sin(60)},-1.2) [whitepiece] () {};

\end{tikzpicture}
\caption{Immediate win and loss features for the White player in Yavalath.}
\label{Fig:WinLossFeaturesYavalath}
\end{figure}

In the game of Yavalath, players win by making a line of four of their pieces, but lose by making a line of three of their pieces beforehand. 
Given these rules, it is relatively easy to construct useful features by hand. For example, \reffigure{Fig:WinLossFeaturesYavalath} depicts three features that activate for moves that result in instant wins or losses. We manually constructed a feature set with only these three features, and manually assigned large weights; $+3000$ for the win-detecting feature, and $-1000$ for each of the loss-detecting features. A \textit{Biased MCTS} player using this feature set throughout complete play-outs achieves a win percentage of $93\%$ against \textit{UCT}, despite a $30\times$ reduction in the MCTS iteration count. This indicates that the poor performance of \textit{Biased MCTS} in Yavalath is not due to a lack of expressiveness in the feature formalisation, but rather due to poor features and/or weights being learned from self-play.

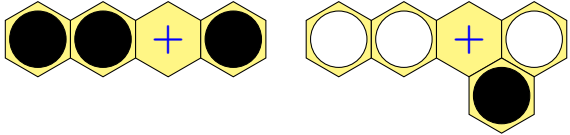
\begin{figure}[t]
\centering
\begin{tikzpicture} [hexa/.style= {shape=regular polygon,regular polygon sides=6,minimum size=1cm,draw,inner sep=0,fill=white!40!yellow,shape border rotate=30},
whitepiece/.style= {shape=circle,minimum size=0.75cm,draw,inner sep=0,fill=white},
blackpiece/.style= {shape=circle,minimum size=0.75cm,draw,inner sep=0,fill=black}]

% first feature cells
\node at (0,0) [hexa] () {};
\node at ({sin(60)}, 0) [hexa] () {};
\node at ({2*sin(60)}, 0) [hexa] () {\Large $\color{blue}{\boldsymbol{+}}$};
\node at ({3*sin(60)}, 0) [hexa] () {};

% first feature pieces
\node at (0,0) [blackpiece] () {};
\node at ({sin(60)},0) [blackpiece] () {};
\node at ({3*sin(60)},0) [blackpiece] () {};

% second feature cells
\node at (4,0) [hexa] () {};
\node at ({4 + sin(60)}, 0) [hexa] () {};
\node at ({4 + 2*sin(60)}, 0) [hexa] () {\Large $\color{blue}{\boldsymbol{+}}$};
\node at ({4 + 3 * sin(60)}, 0) [hexa] () {};
\node at ({4 + 2.5 * sin(60)}, {-0.75}) [hexa] () {};

% second feature pieces
\node at (4,0) [whitepiece] () {};
\node at ({4 + sin(60)},0) [whitepiece] () {};
\node at ({4 + 3 * sin(60)},0) [whitepiece] () {};
\node at ({4 + 2.5 * sin(60)}, {-0.75}) [blackpiece] () {};

\end{tikzpicture}
\caption{Two of the learned Yavalath features with greatest absolute weights (drawn from the perspective of the White player).}
\label{Fig:LearnedFeaturesYavalath}
\end{figure}

\reffigure{Fig:LearnedFeaturesYavalath} depicts two of the most ``important'' features (with large absolute weights) found from self-play in Yavalath. These are easy to understand and seem sensible. The first feature recommends making a move to prevent the opponent from winning in their next turn. The second feature is very similar to the handcrafted win-detecting feature in \reffigure{Fig:WinLossFeaturesYavalath}, with the difference being that it has a seemingly unnecessary requirement for an adjacent opposing piece. This feature can still often detect immediate wins, but not all of them. Despite the poor performance in terms of win percentage in Yavalath, it appears that our proposed approach has still learned sensible -- if not optimal -- features in this game.

\section{Conclusion} \label{Sec:Conclusion}

This paper describes and evaluates an approach for simultaneously learning a set of features, and corresponding weights for a linear policy function, for general games implemented in the \textsc{Ludii} system. The features are formalised in such a way that they are generally applicable, and easily interpretable. The training process is also more easily applicable to general games than, for instance, Deep Neural Networks, which require game-specific knowledge to determine the numbers of input and output nodes a priori.

Using the learned features and weights to bias an MCTS agent, we demonstrate significantly improved game-playing performance over a standard UCT agent in the majority of evaluated board games. This performance is achieved with relatively few self-play games. Out of ten evaluated games, the use of features only reduced playing strength in the game of Yavalath due to computational overhead. Despite the poor performance in this game (which may also lead to poor update targets during self-play), a manual inspection of the top features learned in this game indicates that the approach still discovers sensible features.

In future research, we aim to investigate more approaches for improved feature discovery from self-play. In particular in the game of Yavalath, tests with handcrafted features indicate that it may be useful to pay extra attention to features for endgame positions. Using optimisers with momentum-based terms, rather than a simple Stochastic Gradient Descent optimiser, may enable more rapid learning of large feature weights for features that reliably detect immediate winning or losing moves. We also aim to explore transfer learning between different games, which is already facilitated by the feature representation which is shared across all games, and online fine-tuning of trained policies \cite{Cazenave16PlayoutPolicyAdaptation}.

\section*{Acknowledgment}

This research is part of the European Research Council-funded Digital Ludeme Project (ERC Consolidator Grant \#771292) run by Cameron Browne at Maastricht University's Department of Data Science and Knowledge Engineering. 

\bibliographystyle{IEEEtran}
\bibliography{IEEEabrv,ReferencesAbrv}

\end{document}